\documentclass[runningheads]{llncs}

\usepackage{eccv}

\usepackage{eccvabbrv}
\usepackage{makecell}
\usepackage{tabularx}
\usepackage{graphicx}
\usepackage{booktabs}
\usepackage{marvosym}

\usepackage[accsupp]{axessibility}  

\usepackage{algorithmic}
\usepackage{algorithm}  
\usepackage{float}

\usepackage{array} 
\usepackage{float}

\definecolor{myPink}{rgb}{1,0.9,0.9} 

\usepackage{hyperref}

\usepackage{orcidlink}
\usepackage{multirow} 
\usepackage{color}

\begin{document}

\title{LLM as Dataset Analyst: Subpopulation Structure Discovery with Large Language Model} 

\titlerunning{Subpopulation Structure Discovery with Large Language Model}

\author{Yulin Luo\inst{1*} \and
Ruichuan An\inst{1,2*} \and
Bocheng Zou\inst{1,3} \and \\
Yiming Tang\inst{1,4} \and 
Jiaming Liu\inst{1} \and
Shanghang Zhang\inst{1\dagger}}
\authorrunning{Y. Luo et al.}

\institute{$^1$State Key Laboratory of Multimedia Information Processing, \\School of Computer Science, Peking University \\
$^2$Xi’an Jiaotong University 
$^3$University of Wisconsin-Madison \\
$^4$National University of Singapore\\
\email{yulin@stu.pku.edu.cn}
}

\maketitle


\renewcommand{\thefootnote}{\fnsymbol{footnote}} 
\footnotetext[1]{Equal Contribution.} 
\footnotetext[4]{Corresponding Author.}

\begin{abstract}
The distribution of subpopulations is an important property hidden within a dataset. 
Uncovering and analyzing the subpopulation distribution within datasets provides a comprehensive understanding of the datasets, standing as a powerful tool beneficial to various downstream tasks, including Dataset Subpopulation Organization, Subpopulation Shift, and Slice Discovery.
Despite its importance, there has been no work that systematically explores the subpopulation distribution of datasets to our knowledge. 
To address the limitation and solve all the mentioned tasks in a unified way, we introduce a novel concept of subpopulation structures to represent, analyze, and utilize subpopulation distributions within datasets. 
To characterize the structures in an interpretable manner, we propose the Subpopulation Structure Discovery with Large Language Models (SSD-LLM) framework, 
which employs world knowledge and instruction-following capabilities of Large Language Models (LLMs) to linguistically analyze informative image captions and summarize the structures. 
Furthermore, we propose complete workflows to address downstream tasks, named Task-specific Tuning, showcasing the application of the discovered structure to a spectrum of subpopulation-related tasks, including dataset subpopulation organization, subpopulation shift, and slice discovery. 
With the help of SSD-LLM, we can structuralize the datasets into subpopulation-level automatically, achieve average +3.3\% worst group accuracy gain compared to previous methods on subpopulation shift benchmark Waterbirds, Metashift and Nico++, and also identify more consistent slice topics with a higher model error rate of 3.95\% on slice discovery task for ImageNet. The code will be
available at \url{https://llm-as-dataset-analyst.github.io/}.
\end{abstract}    

\section{Introduction}

\label{sec:intro}
Subpopulation, defined by a set of data points that share common characteristics, is an important concept in machine learning ~\cite{yang2023change}. 
Many tasks are subpopulation-related. For example, image clustering conditioned on text criteria~\cite{kwon2023ictc} is to partition an image dataset into different subpopulations based on user-specified criteria, studying subpopulation shift~\cite{yang2023change, liang2022metashift, zhang2022nico} is to mitigate the negative impact of imbalanced subpopulation distributions in the training set on the model, slice discovery~\cite{eyuboglu2022domino, chen2023hibug} is aimed at identifying subpopulations model underperform.

\begin{figure}[tbp]
    \hsize=\textwidth 
    \centering
    \includegraphics[width=1\linewidth]{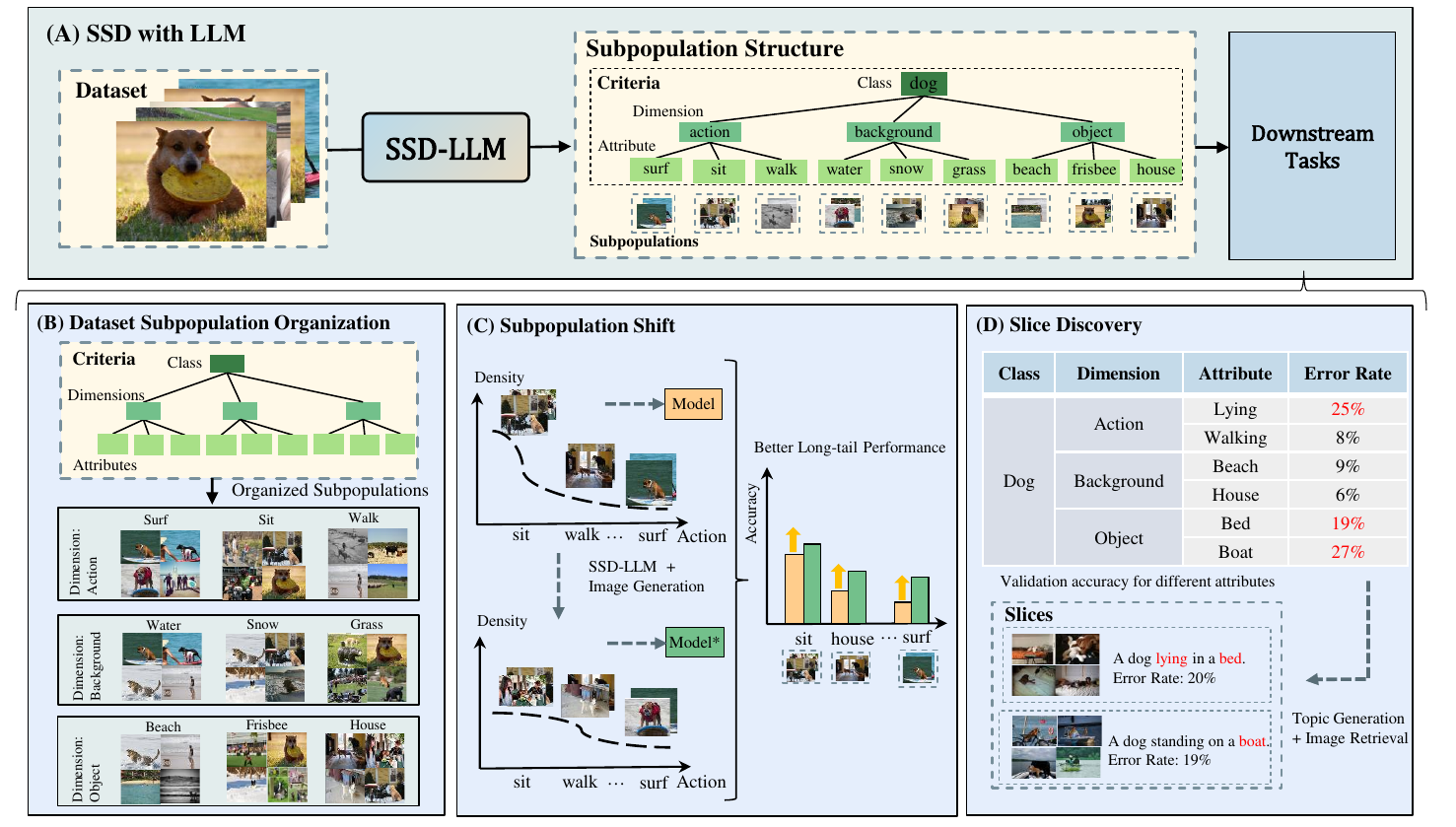}
    \caption{(A) The Workflow of Subpopulation Structure Discovery with Large Language Models (SSD-LLM). SSD-LLM can further support several downstream tasks including: (B) Dataset Subpopulation Organization; (C) Subpopulation Shift; (D) Slice discovery.}
    \label{fig:workflow}
\end{figure}

Summarizing the commonalities of these tasks, we find that analyzing the subpopulation distribution is the key to solving all these problems. If the subpopulation distribution can be characterized, image clustering results under different criteria are naturally obtained~\cite{kwon2023ictc}, additional images can be supplemented to rare subgroups to balance the whole dataset~\cite{dunlap2023diversify}, and slices can be easily discovered by statistics error rate on validation set~\cite{chen2023hibug}.
Despite its importance, existing work~\cite{yang2023change} lacks systematic exploration of subpopulation distribution. 
To adjust the issue, for the first time, we propose the concept of subpopulation structure to represent, analyze, and utilize subpopulation distributions within datasets. 
By definition, a subpopulation structure is a set of hierarchical relations among several subpopulations determined by certain criteria. 

\begin{figure*}[tbp!]
    \centering
    \includegraphics[width=1\linewidth]{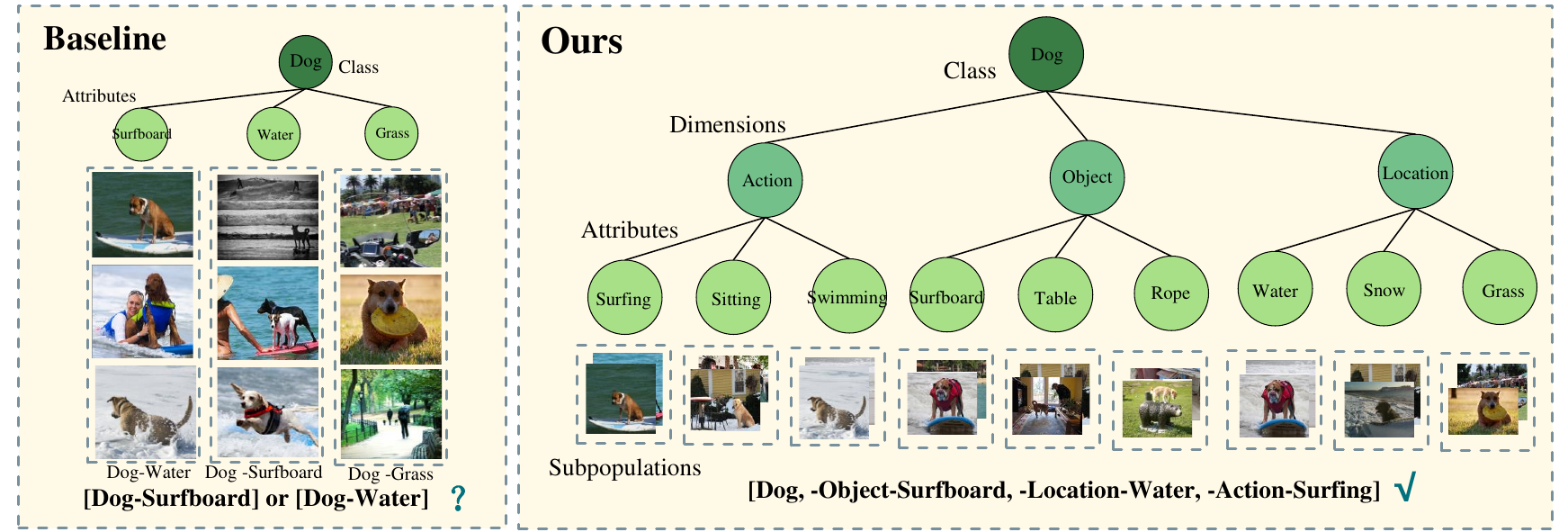}
    \caption{Metashift has the same-level attributes \textit{Surfboard}, \textit{Water}, and \textit{Grass} for class \textit{Dog}, which is irrational due to the possible overlap. As an improvement, we take dimensions into consideration. The class \textit{Dog} has dimensions including \textit{Action}, \textit{Co-occurrence Object}, \textit{Location}, etc., and in dimension \textit{Location}, it includes various attributes like \textit{Water}, \textit{Grass}, etc, which offers a more appropriate assignment for the samples.}
    \label{fig:demo1}
\end{figure*}

Former works like Metashift~\cite{liang2022metashift} and NICO++~\cite{zhang2022nico} have constructed image datasets including the subpopulation information, which organizes the images with respect to some extra attributes, and can be viewed as a class-, attribute-, subpopulation-layer structure. The problem of such a structure is ignoring the category of attributes (or \textit{Dimension}), leading to attribute inconsistency and confusion. To solve this issue, we introduce a class-, dimension-, attribute-, and subpopulation-layer structure. The comparison of the two structures can be seen in Figure. \ref{fig:demo1}. By articulating the classification dimensions, this improved structure provides more nuanced attribute assignments.

Identifying descriptive information within a dataset often requires large amounts of human manufacture~\cite{liang2022metashift,zhang2022nico}, which urges the need of automatic workflows to complete the subpopulation structure discovery. However, automatically identifying subpopulation structures within image datasets presents a significant challenge. The approach must be capable of extracting key information from images and summarizing essential content from extensive texts. Furthermore, it necessitates comprehensive world knowledge, enabling a broad understanding of various aspects of the datasets, including diverse categories, common attributes, and the relationships between dimensions and attributes.

Recently, Large Language Model (LLM)~\cite{liu2023chatgpt, kojima2023large, Wei2022ChainOT}  and Multimodal Large Language Model (MLLM)~\cite{liu2023visual, gao2023llama} have attracted wide attention due to their superior capacities. LLM has shown extensive world knowledge and remarkable abilities in summarization, instruction following~\cite{liu2023chatgpt}, etc. MLLM extends the capabilities of LLM to handle visual inputs. By visual instruction tuning~\cite{liu2023visual}, MLLM can verbalize the rich information of images. Motivated by these, we propose a novel framework \textbf{Subpopulation Structure Discovery with Large Language Model(SSD-LLM)}, illustrated in Figure. \ref{fig:workflow}, to automatically uncover the structure. The core idea is to generate informative captions from images with MLLM, followed by analyzing and summarizing the subpopulation structure of datasets with LLM. Specifically, we design two elaborate prompt engineering components, \textbf{Criteria Initialization} and \textbf{Criteria Refinement}. The former utilizes a generate-and-select paradigm to summarize dimensions and attributes sequentially. The latter employs self-consistency as an indicator to evaluate and refine the criteria.
After obtaining complete criteria, each image is assigned to corresponding attributes according to its caption. The final subpopulation structures can be leveraged to finish various downstream tasks with the help of our proposed \textbf{Task-specific Tuning}. In this work, we focus on three application scenarios, i.e. dataset subpopulation organization, subpopulation shift, and slice discovery. 
We validate the effectiveness of SSD-LLM on these subpopulation-related tasks. For subpopulation shift, we achieve an improvement of \textbf{+3.3\%} in worst group accuracy across three datasets compared to SOTA methods, and for slice discovery, we can identify more consistent slice topics with a higher model error rate of \textbf{3.95\%}. 

Our contributions are summarized as follows:
\begin{itemize}
    \item We introduce the concept of subpopulation structure to characterize subpopulation distribution in an interpretable manner for the first time.
    \item We propose class-dimension-attribute-subpopulation structure, reducing the attribute confusion of the current class-attribute-subpopulation structure.
    \item We propose Subpopulation Structure Discovery with Large Language Model (SSD-LLM) framework to uncover the underlying subpopulation structure of datasets automatically, with two elaborate prompt engineering components Criteria Initialization and Criteria Refinement.
    \item We provide methods for Task-specific Tuning, enabling the application of the structures across a spectrum of subpopulation-related tasks, including dataset subpopulation organization, subpopulation shift, and slice discovery. 
\end{itemize}
\section{Related Works}
\label{sec:related_work}

\subsection{Hierarchical Structure of Image Datasets }
Recent research has emphasized the need to organize datasets into hierarchical structures allowing for benchmarking various downstream tasks~\cite{liang2022metashift, zhang2022nico, WahCUB_200_2011, russakovsky2015imagenet}.
Metashift~\cite{liang2022metashift} builds a collection of 12,868 sets of images related to 410 main subjects and their contexts. NICO++~\cite{zhang2022nico}, Waterbirds~\cite{WahCUB_200_2011}, and ImageNetBG~\cite{russakovsky2015imagenet} also propose methods for constructing various types of hierarchical datasets.
However, the construction of these hierarchical datasets often requires manual annotation, hindering automatic construction. These approaches focus on just a single dimension, such as object context in Metashift, background in Waterbirds, and ImageNetBG, while more practical scenarios may involve multiple dimensions hidden within the comprehensive visual information.

\subsection{Extract Information from Image Captions}
Recent works such as ALIA \cite{dunlap2023diversify},  VeCAF\cite{zhang2024vecaf}, Bias2Text \cite{kim2023biastotext}, and ICTC \cite{kwon2023image} explore utilizing caption models to obtain information from datasets. ALIA provides a method to augment datasets by generating variations of existing images through captioning and text-to-image models. While ALIA\cite{dunlap2023diversify} supports dataset improvement, it lacks knowledge about attribute types, bias, or subpopulation shift existence. Bias2Text\cite{kim2023biastotext} detects dataset bias by transforming images into descriptive captions and keywords. However, without large language model participation, Bias2Text fails to support classification dimension selection and can only differentiate images with basic keywords. More recently, ICTC\cite{kwon2023image} enables conditional image clustering using an LLM in a straightforward manner. Although ICTC clusters images when given the criterion, it requires human-assigned text prompts. Compared to ICTC, our prompt engineering paradigm supports scalable automatic subpopulation structure dataset organization without human criteria assignment and can generate comprehensive criteria tailored to datasets for various downstream tasks.

\subsection{LLM Prompt Engineering}
As the popularity of LLMs has surged, prompt engineering, the process of crafting and refining prompts to guide LLMs towards desired outputs\cite{kojima2023large, Wei2022ChainOT} has also shown more and more importance. Various prompt engineering methods\cite{liu2021pretrain, qiao2023reasoning} and principles have emerged, and researchers or engineers have explored their applications in a diverse range of downstream tasks\cite{cui2023chatlaw,qian2023communicative,liu2023chatgpt,park2023generative,fu2023improving}.
In particular, in-context learning has emerged as a pivotal technique, validated both experimentally\cite{dong2023survey, li2023mot} and theoretically\cite{luo2023prompt, dai2023gpt, xie2022explanation}. This approach involves providing the LLM with context information relevant to the task at hand, enabling it to generate more accurate and relevant responses.
Least-to-most prompting\cite{zhou2023least} breaks down complex tasks into smaller, more manageable steps, enhancing LLM's reasoning skill by querying the LLM with more simplified sub-questions.
Self-consistency\cite{Wang2022SelfConsistencyIC} proposes to ensemble multiple responses to the LLM given the same prompt to get enhanced results, suggesting that consistent responses as an indicator for correct problem solving.
Self-refining\cite{madaan2023selfrefine} demonstrates that we can use LLMs to refine their outputs by themselves with careful designing of prompts.
In this work, we leverage a combination of prompt engineering techniques, including in-context learning, chain-of-thought, self-consistency, and self-refining to tackle subpopulation structure discovery effectively.
\section{Method}
In this section, we introduce our proposed method, subpopulation structure discovery with large language models(SSD-LLM). We describe the overall pipeline of SSD-LLM in Section \ref{overview}, outlining how the paradigm automatically discovers the latent subpopulation structures inside the dataset. The process begins with captioning the images in the dataset with an MLLM, detailed in Section \ref{extraction}, and proceeds with Criteria Initialization with an LLM in Section \ref{initialization}. The paradigm then refines the initialized criteria through a recursive self-refinement procedure, detailed in Section \ref{refinement}. Finally, images are assigned to attributes, completing the subpopulation structure discovery process, as elaborated in Section \ref{assignment}. The section concludes with a discussion of how to apply the method to various downstream tasks, as presented in Section \ref{tuning}.

\begin{figure*}[tbp]
    \centering
    \includegraphics[width=1\linewidth]{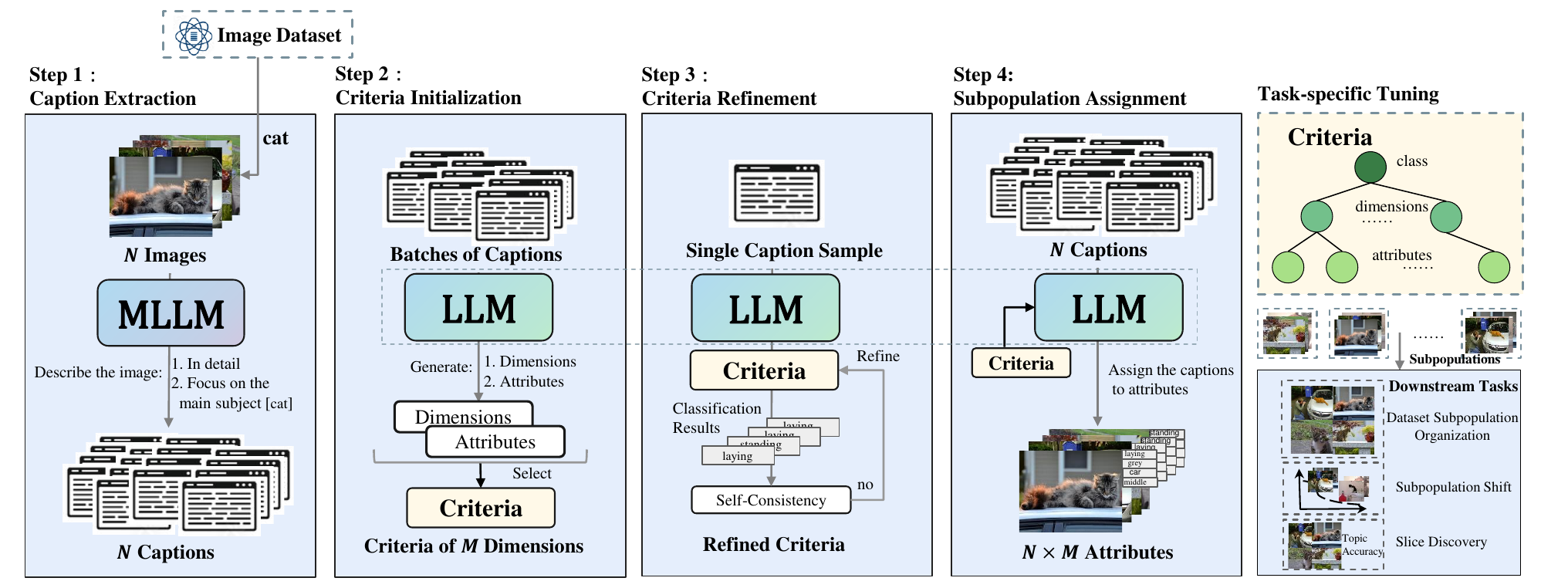}
    \caption{Subpopulation Structure Discovery with Large Language Model (SSD-LLM). (Step 1) Multimodality Large Language Model (MLLM) extracts informative captions from images. (Step 2) LLM initializes the criteria with a sample-based generate-and-select paradigm. (Step 3) LLM refines the criteria using self-consistency as an indicator. (Step 4) LLM assigns each caption with specific attributes according to the refined criteria, uncovering the intrinsic subpopulation structures hidden in the dataset. The resulting criteria and subpopulations are used in several downstream tasks.}
    \label{fig:pipeline}
\end{figure*}

\subsection{Overview}
\label{overview}
To automatically discover the subpopulation structures, 
we propose a novel prompt engineering paradigm that effectively leverages the capabilities of both multimodal large language models (MLLMs) and large language models (LLMs).
Our proposed method comprises four key steps (Figure. \ref{fig:pipeline}). First, we transform the images into information-rich captions that capture the main information in the images using MLLMs. Second, we employ a novel sample-based prompt engineering method to guide an LLM to produce criteria consisting of dimensions and corresponding attributes organizing the dataset. Third, we prompt the LLM to refine this generated criteria. Last, we assign all the images in the dataset to specific attributes accordingly, uncovering the intrinsic subpopulation structures in the dataset, and paving the way for further analysis about the dataset. Detailed descriptions of each step are provided below. 
For notations, consistent with various former works~\cite{luo2023prompt,Wei2022ChainOT}, we denote the operation of getting responses from the language models as $LLM$ and $MLLM$, and use $[,]$ to represent the concatenation operation of two pacts of texts.

\subsection{Caption Extraction}
\label{extraction}
To begin our approach, we leverage the powerful image captioning capabilities of the MLLM to transform the images into informative and detailed captions. Instead of briefly describing the images,  we prompt the MLLM to generate more detailed captions centered around the main subject {CLS}. 
The prompt we used in this step is stated as follow:

$P_1$ = \textit{"Describe the image of the subject CLS in detail."}

\begin{tabular}{p{0.45\textwidth}p{0.45\textwidth}}
\begin{minipage}{.45\textwidth}
\begin{algorithm}[H]
\label{alg: step1}
\caption*{\textbf{Step 1} Caption Extraction}
\begin{algorithmic}[1]
\REQUIRE Dataset: $D_{img}$, MLLM
\ENSURE Image Captions: $C$
\STATE \textbf{for} i \textbf{in range}(NumOfIterations)
\STATE \quad\indent $img$ = $D_{img}$.sample()
\STATE \quad\indent $c$ = MLLM($img$, $P_1$)
\STATE \quad\indent $C$.append($c$)
\STATE \textbf{end for}
\end{algorithmic}
\end{algorithm}
\end{minipage}
\end{tabular}
\begin{minipage}{.45\textwidth}
\begin{algorithm}[H]
\label{alg: step2-1}
\caption*{\textbf{Step 2-1} Dimension Generation}
\begin{algorithmic}[1]
\REQUIRE  Captions: $C$, LLM
\ENSURE  dimensions: $Dims$
\STATE \textbf{for} i \textbf{in range}(NumOfIterations)
\STATE \quad\indent $c$ = $C$.sample(NumOfSamples)
\STATE \quad\indent $S$.append(LLM([$P_2^1$,$c$]))
\STATE \textbf{end for}
\STATE $Dims$ = LLMSummary($S$)
\end{algorithmic}
\end{algorithm}
\end{minipage}

\subsection{Criteria Initialization}
\label{initialization}
To discover the hidden subpopulation structures within the dataset, we employ an LLM to delve into the information-rich captions generated in the previous step. Our objective is to identify certain criteria that effectively partition the images into several distinct subgroups. Beyond simply dividing the dataset into subgroups, we articulate the classification dimension for the partition and record all the resulting attributes generated from the classification process. Along with the class information and the resulting subpopulations, this criteria naturally form a four-layer structure, class-, dimension-, attribute-, and subpopulation-. Noticing criteria encompass multiple dimensions and their corresponding attributes, we adopt a generate-and-select paradigm with the LLM to discover the dimensions and the attributes sequentially.

To determine the dimensions and attributes, we employ an iterative sampling approach, repeatedly prompting the LLM to propose dimensions and attributes based on batches of image captions. In each iteration, the LLM generates candidate dimensions and attributes, which are subsequently processed through an LLM summarization process. This sample-and-summarize approach effectively addresses the challenges when processing large datasets. 
Since the number of dimensions that can differentiate images in a dataset is relatively small, and these dimensions have an appearance in numerous images, our sample-based approach effectively identifies relevant dimensions. 
The prompts we used in this step are stated as follow, omitting Chain-of-thought examples for simplicity.

$P_2^1$ = \textit{"Suggest some dimensions that can differentiate the following image captions."}

$P_2^2$ = \textit{"Suggest a complete criterion to differentiate the following image captions by the given dimension."}

\subsection{Criteria Refinement}
\label{refinement}
To further refine the criteria and ensure its effectiveness in classifying image captions across the dataset, we implement a recursive refining process. This approach proposes a novel method for identifying image captions requiring further refinement utilizing the \textbf{self-consistency} of LLM responses as an indicator~\cite{Wang2022SelfConsistencyIC}. 
\begin{algorithm}[H]
\caption*{\textbf{Step 2-2} Attribute Generation}
\begin{algorithmic}[1]
\REQUIRE  Dimensions: $Dims$, Captions: $C$, Large language model: LLM
\ENSURE  Initialized criteria: $Criteria$
\STATE \textbf{for} $dim$ \textbf{in} $Dims$ \textbf{do}
\STATE \quad\textbf{for} i \textbf{in range}(NumOfIterations) \textbf{do}
\STATE \quad\quad\indent $c$ = $C$.sample(NumOfSamples)
\STATE \quad\quad\indent $S$.append(LLM([$P_2^2$, $dim$, $c$]))
\STATE \quad\textbf{end for}
\STATE \quad$Attributes$ = LLMSumarry($S$) *list of attributes
\STATE \quad$Criteria[dim]=Attributes$
\STATE \quad$S$.reset()
\STATE \textbf{end for}
\end{algorithmic}
\end{algorithm}
This choice stems from our empirical observation that if an image can be accurately classified according to a particular dimension, it should consistently be classified into the same attribute multiple times. Inconsistent responses, however, suggest that the current criteria require further refinement, either to merge redundant attributes or to include new attributes. 
The prompts we used in this step are stated as follow and the pseudocode is included in the appendix:

$P_3^1$ = \textit{"Classify the caption by the criteria listed below."}

$P_3^2$ = \textit{"We are unable to classify the following image caption using the provided criteria due to attributes redundancy or misappearance. If redundancy, please prune the redundant attributes. If missing, please suggest an additional attribute that would match the image caption."}


\subsection{Subpopulation Assignment}
\label{assignment}
Equipped with the comprehensive criteria, we proceed to systematically assign each image to the specific attributes of each dimension. Images assigned to the same attributes across all dimensions form distinct subgroups within the dataset, revealing the intrinsic subpopulation structures hidden within the data. These subpopulation structures can then be leveraged to perform various downstream tasks, completing our overall pipeline for employing an LLM to analyze the image dataset.
The prompt we used in this step is stated as follow:

$P_4$ = \textit{"Please assign following caption to one attribute of given dimension."}
\begin{algorithm}[h]
\label{alg: step4}
\caption*{\textbf{Step 4} Subpopulation Assignment}
\begin{algorithmic}[1]
\REQUIRE  Captions: $C$, Large language model: LLM, criteria: $Criteria$
\ENSURE  Further assignments for each caption $c$.
\STATE \textbf{for} $c$ \textbf{in} $C$ \textbf{do}
\STATE \quad\textbf{for} [$dim$, $Attributes$] \textbf{in} $Criteria$ \textbf{do}
\STATE \quad\quad\indent $c$.assign(LLM([$P_4$,$c$,$dim$,$Attributes$]))
\STATE \quad\textbf{end for}
\STATE \textbf{end for}
\end{algorithmic}
\end{algorithm}

\begin{figure*}[tbp]
    \centering
    \includegraphics[width=1\linewidth]{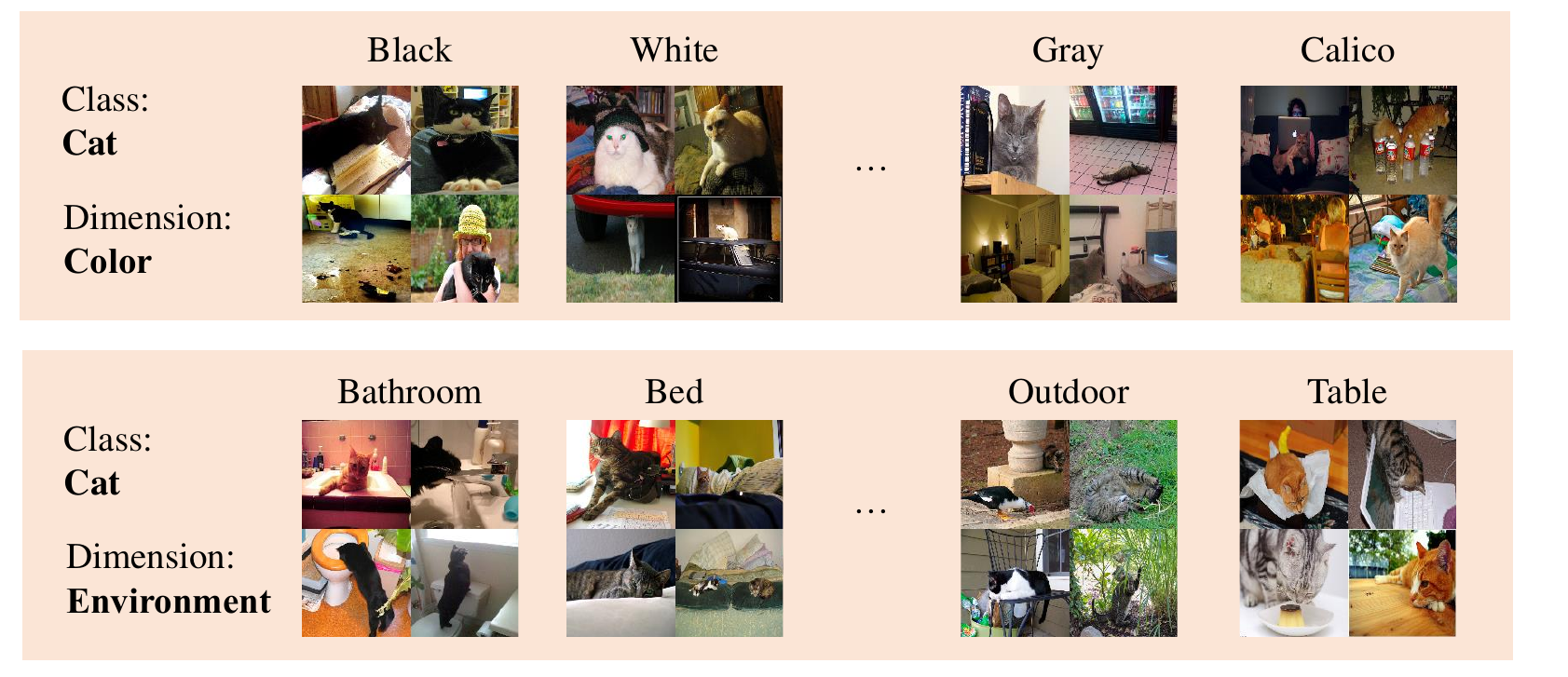}
    \caption{A visualization of organised subpopulations in a dataset of cats.}
    \label{fig:experiment11}
\end{figure*}

\subsection{Task-specific Tuning for Downstream Tasks}
\label{tuning}
Once we have identified the criteria and subpopulations within the dataset, we can leverage this information to tackle several downstream tasks effectively. This includes organizing the subpopulations, which can reveal valuable insights into the data, such as potential subpopulation biases and the presence of long-tail attributes. In practice, the subpopulation structures can be used to improve model performance on various tasks if combined with extra operations, including handling subpopulation shifts and slice discovery.

\noindent {\textbf{Dataset Subpopulation Organization} }
Organizing subpopulations within a dataset can provide beneficial information about the numbers of data points in each subpopulation and can reflect dataset biases and help us identify some long-tail subgroups. 
We conduct experiments on the task of image clustering conditioned on human-specified criteria to evaluate the quality of subpopulation organization.
In specific, when organising the subpopulations of a given image dataset, we first select out the relevant dimensions and then attach attributes assigned by SSD-LLM directly to the images accordingly.
Our method, SSD-LLM, automates subpopulation organization within datasets, and thus holds the potential to revolutionize the way hierarchical datasets are constructed.

\noindent {\textbf{Subpopulation Shift}}
Subpopulation shift~\cite{yang2023change} stands as a common challenge in machine learning, occuring when the proportion of some subpopulations between training and deployment changes, and is shown to be of significant influence to model performances~\cite{yang2023change, santurkar2020breeds}.
SSD-LLM, combined with image generation, offers a solution to better handle the scenarios of subpopulation shifts. In our experiments, after we apply SSD-LLM to the datasets, we collect statistics the number of images contained in each subpopulation and utilize diffusion model to generate images for underrepresented subpopulations. Subsequently, we sample attributes from the subpopulation structure for each underrepresented subpopulation and employ LLM to make complete sentences based on these words as the input prompt of a diffusion model. The diffusion model generates images augmented to the image dataset which further helps to achieve balanced classes and attributes. 
Moreover, we propose to harness an LLM to suggest extra dimensions and attributes based on the current sets in this task for enriched subpopulation structure, generating more diverse images. 

\noindent {\textbf{Slice Discovery}}
Slice discovery is a task aiming at uncovering subpopulations within a dataset where a machine learning model consistently exhibits poor performances. 
These subpopulations with underperformances, or slices, provide valuable insights into the model's limitations, potential biases, and how to improve the performances.
SSD-LLM conducts slice discovery for an image dataset with the help of the assigned attributes. In detail, we first calculate the error rates on all subpopulations discovered with the SSD-LLM. Then we identify out the subpopulations with the highest error rate and use the LLM to summarize out discriptions based on the attributes of the subpopulations in the form of texts representing the slice topics, completing the task of slice discovery.

\section{Experiments}
We now present experimental results demonstrating the effectiveness of SSD-LLM. In particular, we present the main settings and results in this section and defer extra details, including various visualization result, to appendix. In our experiments,
we mainly use LLaVA1.5~\cite{liu2024visual} for MLLM and GPT-4~\cite{achiam2023gpt} for LLM.

We conduct experiments on Dataset Subpopulation Organization, Subpopulation Shift, and Slice Discovery. Our superior performance underscores the method's efficacy in identifying and analyzing subgroups, further affirming its utility in addressing related challenges. Moreover, it illustrates that the unified paradigm can effectively address a variety of downstream tasks.



\newcommand\MyXhilneF[0]{\Xhline{2.5\arrayrulewidth}}
\renewcommand\arraystretch{1}
\setlength\tabcolsep{3pt}
\begin{table}[!t]
    \footnotesize
    \centering
\begin{tabular}{ccccc}
\toprule
Dataset & Criterion & SCAN* & \makecell{IC|TC} & Ours \\
\midrule
\multirow{3}{*}{\makecell{Stanford \\40 Action}}
 & Action & 0.346 & 0.747 & \textbf{0.817}\\
 & Location & 0.357& 0.671& \textbf{0.705}\\
& Mood & 0.276 &0.746 & \textbf{0.768} \\
\hline
Place365 & Place & 0.332 & - &\textbf{0.696} \\
\hline 
PPMI & Musical Instruction & 0.598 & 0.934 & \textbf{0.955} \\
\hline
Cifar10 & Object & 0.839 & 0.911 & \textbf{0.921} \\
\hline
STL10 & Object & 0.798 & 0.986 & \textbf{0.988} \\
\bottomrule
\end{tabular}
    \caption{Quantitative results of Dataset Subpopulation Organization.  Method labeled with * is evaluated by having a human provided
ground truth labels, cause the method itself is an unsupervised learning paradigm.}
    \label{tab:assign}
\end{table}

\subsection{Evaluation on Dataset Subpopulation Organization}
\label{subpopulation organization}


\noindent {\textbf{Setup}} \label{setup4}
The data assignment process facilitated by SSD-LLM, can be considered a form of clustering. To evaluate the quality of these identified subpopulations, we assess the clustering accuracy by comparing images against secondary labels that reflect subgroup attributes derived via SSD-LLM. 
Full information for datasets, text criterion, and model selection can be found in appendix.


\noindent {\textbf{Comparison Methods}}
SCAN~\cite{van2020scan} is a two-stage clustering method that decouples feature learning and clustering. 
IC|TC~\cite{kwon2023image} is a new paradigm for image clustering that supports human interaction. It utilizes the given Text Criteria to accurately control the quality of the clustering results. 



\noindent {\textbf{Results and Analysis}}
In Table.~\ref{tab:assign}, we report the average accuracy achieved by each method based on the predefined textual criteria. When compared to IC|TC, SSD-LLM demonstrates competitive performance. It is important to note that IC|TC incorporate artificial judgment in its process, which leads to poor scalability when handling large datasets. In contrast, our approach is fully automated. 
We conduct visualization of our organized subpopulations, as shown in Figure.~\ref{fig:experiment11}. We can visually observe that the pictures and dimensions are indeed consistent, indicating the effectiveness of our mining and assigning process.



\newcommand\MyXhlineC[0]{\Xhline{2.5\arrayrulewidth}}
\renewcommand\arraystretch{1.0}
\setlength\tabcolsep{2pt}

\begin{table*}[!t]
    \centering
    \resizebox{.98\linewidth}{!}{%
        \begin{tabular}{c|c|c|c|c|c|c|c|c|c}
            \MyXhlineC
            {\multirow{2}*{Type}} & {\multirow{2}*{Method}} & \multicolumn{4}{c|}{Average Accuracy} & \multicolumn{4}{c}{Worst Group Accuracy} \\
            \cline{3-10}
            \multicolumn{1}{c|}{}&\multicolumn{1}{c|}{} &  Waterbirds & Metashift & Nico++ & Average & Waterbirds & Metashift& Nico++ & Average \\
            \MyXhlineC
            \multirow{1}{*}{Vanilla} & ERM & 84.1 & 91.2& 76.3& 83.7 & 69.1 & 82.1 &17.8 & 56.3\\
            \cline{1-10}
            \multirow{4}{*}{\makecell{Subgroup Robust\\ Methods}} & GroupDRO & 86.9 & 91.5 & 74.0& 84.1& 73.1 & 83.1 &12.2 &56.1 \\
            & JTT & 88.9 & 91.2 &77.5 &85.9& 71.2 & 82.6 & 15.6& 56.5\\
            & LfF & 86.6 & 80.4 & 77.5& 81.5& 75.0 & 72.3 &15.6 &54.3 \\
            & LISA & 89.2 & 91.4 & 75.0&85.2& 77.0 & 79.0 & 18.9& 58.3\\
            \cline{1-10}
            \multirow{5}{*}{\makecell{Imbalanced\\ Learning}} & Resample & 86.2 & \underline{92.2} &77.3 &85.2 & 70.0 & 81.0 & 16.7&55.9 \\
            & Reweight & 86.2 & 91.5 & 73.8& 83.8& 71.9 & 83.1 &12.2 & 55.7\\
            & Focal & \underline{89.3} & 91.6 &73.1 &84.7 & 71.6 & 81.0 &16.7 &56.4\\
            & CBLoss & 86.8 & 91.4 & 76.3&84.8 & 74.4 & 83.1 &12.2& 56.6 \\
            & BSoftmax & 88.4 & 91.3 &74.2 &84.6& 74.1 & 82.6 & 16.7&57.8 \\
            \cline{1-10}
            \multirow{2}{*}{\makecell{Traditional Data\\ Augmentation}} & Mixup & 89.2 & 91.4 &73.0 &84.5 & \underline{77.5} & 79.0 &14.4 & 57.0\\
            & RandAug & 86.3 & 90.9 &72.0 &83.1 & 71.4 & 80.9 &16.7 &56.3 \\
            \cline{1-10}
            \multirow{3}{*}{Diffusion} & Class Prompt & 85.9 & 91.5 & 78.0 & 85.1 & 71.3 & 82.7 & 18.5 & 57.5 \\
            & \makecell{Class-Attribute} & 89.1 & 91.4 & \underline{78.6} & \underline{86.4} & 73.5 & \underline{83.8} & 18.8 & \underline{58.7} \\
            & CiP & 88.0 & 91.1 & 78.3 & 85.8 & 73.5 & 82.4 & \underline{19.3} & 58.4 \\
            
            \cline{1-10}
            \multirow{1}{*}{LLM+Diffusion}  & \textbf{SSD-LLM (Ours)} & \textbf{90.5} & \textbf{93.0} & \textbf{80.4}&\textbf{88.0} & \textbf{79.1} & \textbf{84.8} & \textbf{22.1} &\textbf{62.0}  \\
            \MyXhlineC
        \end{tabular}
    }
    \caption{Comparison of methods for image classification with subpopulation shifts.}
    \label{real_results}
\end{table*} 

\subsection{Evaluation on Subpopulation Shift}\label{pred}



\noindent {\textbf{Setup}} We evaluate subpopulation shifts on three commonly used image datasets, Metashift (Cats vs Dogs), Waterbirds (Landbirds vs Waterbirds), and NICO++. We choose \textit{Average Accuracy} and \textit{Worst Group Accuracy} as evaluation metrics. 
To ensure a fair comparison, following~\cite{yang2023change}, we conduct a random search of 16 trials over a joint distribution of all hyperparameters. We then use the validation set to select the best hyperparameters for each algorithm, fix them, and rerun the experiments under five different random seeds to report the final average results. To make the evaluation more realistic, we consider the model selection setting \textit{Attributes are unknown in both training and validation}. 

\noindent {\textbf{Comparison Methods}}
Following recent benchmarking efforts~\cite{yang2023change}, we compare SSD-LLM with several types of methods: (1) \textit{vanilla:} ERM~\cite{vapnik1999overview}, (2) \textit{Subgroup Robust Methods:} GroupDRO~\cite{sagawa2019distributionally}, LfF~\cite{nam2020learning}, JTT~\cite{liu2021just}, LISA~\cite{yao2022improving}, (3) \textit{Imbalanced Learning:} ReSample~\cite{japkowicz2000class}, ReWeight~\cite{japkowicz2000class}, Focal~\cite{lin2017focal}, CBLoss~\cite{cui2019class}, Bsoftmax~\cite{ren2020balanced}, (4) \textit{Traditional Data Augmentation:} Mixup~\cite{zhang2017mixup}, RandAug~\cite{cubuk2020randaugment}, (5) \textit{Diffusion:} Class Prompt~\cite{shipard2023diversity}, Class-Attribute Prompt~\cite{shipard2023diversity}, CiP~\cite{lei2023cip}. 

\noindent {\textbf{Results and Analysis}}
From Table \ref{real_results}, SSD-LLM surpasses previous methods, with a +1.6\% improvement in average accuracy and +3.3\% in worst group accuracy across three datasets. The analysis is as follows: (1) Despite being based on conventional ERM, data-based approaches show competitive performance compared to model-based algorithms, highlighting their potential. (2) For diffusion-based methods, class-attribute prompts outperform class prompts, underscoring the importance of understanding dataset imbalanced attributes for effective image generation. However, the need for pre-identifying these attributes emphasizes the value of SSD-LLM, which automates attribute discovery and provides detailed annotations, enhancing performance. (3) The superior performance of CiP over Class Prompt underscores the significance of diverse text prompts. (4) Our method's superior performance results from a comprehensive analysis of subpopulation imbalances within the dataset. The strategic text diversity achieved through attribute sampling and LLM sentence-making effectively addresses subpopulation shifts.

\renewcommand\arraystretch{1}
\setlength\tabcolsep{3pt}
\begin{table}[!t]
    \footnotesize
    \centering
    \begin{tabular}{c|cccccc|c}
\hline
\multicolumn{1}{c|}{Method|Categories} & Boat & Bird & Car & Cat & Dog & Truck & \makecell{Topic Error\\Rate} \\
\hline
ImageNet & 4.33 & 0.81 & 11.33 & 11.14 & 0.69 & 11.71 &6.72  \\
General Prompt & 47.82 & 12.11 & 43.55 & 14.22 & 10.19 & 12.65 & 23.42 \\
GPT-Suggest & 57.55 & 12.87 & 43.59 & 12.71 & 16.34 & 28.12 & 28.53 \\
Domino(Bert) & 76.26 & \underline{42.26} & 54.21 & 33.89 & \underline{24.50} & 29.54 & \underline{43.44} \\
B2T & \underline{77.62} & 30.04 & \underline{58.17} & \textbf{36.36} & 19.80 & \underline{33.47} &  42.58\\
SSD-LLM (Ours) & \textbf{79.31} & \textbf{45.67} & \textbf{60.34} &  \underline{32.97} & \textbf{26.48} & \textbf{39.57} & \textbf{47.39} \\
\hline
\end{tabular}
    \caption{Results of slice discovery on Imagenet-1K with various SDMs.}
    \label{tab:slice}
\end{table}

\subsection{Evaluation on  Slice Discovery}
 
 
\noindent{\textbf{Setup}} \label{dominosetup}
In contrast to typical slice discovery tasks, we redefine  evaluation pipeline following~\cite{gao2023adaptive}. In this study, we evaluate bugs found of ImageNet models.
Specifically, a classification is deemed incorrect when an image containing target object is erroneously identified by the model as containing an unrelated object.

\noindent{\textbf{Comparison Methods}} \label{comp}
Domino~\cite{eyuboglu2022domino} represents a state-of-the-art method in slice discovery, which effectively clusters errors identified in the validation set and characterizes them through captions generated automatically. B2T~\cite{kim2023bias} is a recently proposed framework which identifies and interprets visual biases in vision models using keyword extraction from captions of mispredicted images.

\noindent{\textbf{Results and Analysis}}
We evaluate the effectiveness of our method in slice discovery on 6 representative superclasses in Imagenet. As shown in Table. ~\ref{tab:slice}, our SSD-LLM overcomes all other SDMs, including Domino~\cite{eyuboglu2022domino}, by a significant margin. We find topics given by Domino tend to encounter two unsatisfactory cases: loss of semantics, and missing class information. 
These cases are also discussed in~\cite{gao2023adaptive}, where they reasoned this phenomenon into the inherent difficulties of automatic SDMs. However, evidence suggests that our SSD-LLM can handle these problems, while keeping a high error rate and maintaining automation. Specifically, SSD-LLM achieves an average error rate of  
\textbf{47.39\%}, surpassing Domino~\cite{eyuboglu2022domino} by \textbf{3.95\%}. Furthermore, when we trace back to origin dataset, the discovered slice is also very consistent(detailed visualizations included in appendix). Interestingly, we find the data mining process of SSD-LLM is just the same as human data scientists, who take up hypotheses and improve model performance by viewing batches of bad subpopulations.
Experiments show that our method overcome the inherent difficulties while maintaining automaton, paving the way for data-centric methods.

\subsection{Ablation Study}

\nonumber \textbf{{Hyperparameters of Criteria Initialization/Refinement}}
The N $\times$ M in Table. \ref{ablation_criteria} represents the rounds and samples of suggestions. Setting A serves as our baseline. For A, B, and C, results show that insufficient suggested dimensions lead to decreased performance, while enough dimension samples lead to stable performance, as the ICTC task requires only the most appropriate match. Too few suggestions may fail to find suitable dimensions, resulting in mismatched attribute generation. For A, D, and E, results highlight the importance of the number of captions in summarizing attributes. Too many captions can cause some to be overlooked, reducing total identified attributes and performance. The improvement from A to F demonstrates that Criteria Refinement enhances attribute comprehensiveness and final performance.


\begin{table}[t!]
\centering
\begin{tabular}{|c|c|c|c|c|c|c|c|}
\hline
\multirow{2}{*}{Methods} & \multirow{2}{*}{Index} & \multicolumn{2}{c|}{Criteria Initialization} & \multirow{2}{*}{\makecell{Criteria\\ Refinement} }& \multicolumn{3}{c|}{Accuracy} \\
\cline{3-4}
\cline{6-8}
  & & Dimension & Attribute & & Action & Location & Mood \\
\hline
\multirow{6}{*}{\makecell{SSD\\LLM}} & A & 10 $\times$ 20 & 20 $\times$ 20 & $\times$ & 75.6 & 67.3 & 71.6 \\
\cline{2-8}
 & B & 5 $\times$ 20 & 20 $\times$ 20 & $\times$ & 65.2 & 60.0 & 66.2 \\
\cline{2-8}
 & C & 20 $\times$ 20 & 20 $\times$ 20 & $\times$ & 75.9 & 67.1 & 71.3 \\
\cline{2-8}
 & D & 10 $\times$ 20 & 10 $\times$ 40 & $\times$ & 70.3 & 68.9 & 69.5 \\
\cline{2-8}
 & E & 10 $\times$ 20 & 40 $\times$ 10 & $\times$ & 80.2 & 70.0 & 74.4 \\
\cline{2-8}
 & F & 10 $\times$ 20 & 20 $\times$ 20 & \checkmark & 81.7 & 70.5 & 76.8\\
\hline
\end{tabular}
\caption{Ablation study on the number of samples (NUM of rounds$\times$NUM per round), and Criteria Refinement (with or without).}
\label{ablation_criteria}
\end{table}

\noindent \textbf{Diffusion Generation Strategy}
From Table. \ref{ablation_diffusion}, we analyze key components for employing diffusion models to address subpopulation shift. For A$\rightarrow$B, we confirm that managing imbalance attributes within datasets helps solve the task. For B$\rightarrow$D, we illustrate that a comprehensive subpopulation structure benefits the task. For D$\rightarrow$E, balanced subpopulation sampling improves data quality and model training. For E$\rightarrow$F, LLM-generated prompts produce more reasonable images, enhancing results. For F$\rightarrow$G, LLM suggests additional attributes to enrich subpopulation structures, generating more diverse images and improving model generalization. For F$\rightarrow$H, we verify the scaling capability of our SSD-LLM.

\begin{table}[t!]
\scalebox{0.73}{
\begin{tabular}{|c|c|c|c|c|c|c|c|}
\hline
Methods                  & Index & Attribute Mode      & Sample Mode & \makecell{SD Prompt\\ Mode} & Number & \makecell{Average\\ Accuracy} & \makecell{Worst Group\\ Accuracy} \\ \hline
ERM                      & ——    & ——                   & ——          & ——              & $\times$1      & 84.1             & 69.1                 \\ \hline
\multirow{8}{*}{SSD-LLM} & A     & GT Attribute Unknown & random      & Direct          & $\times$1     &   85.9               &      71.3                \\ \cline{2-8} 
 & B     & GT Attribute Known   & random      & Direct          & $\times$1     &   89.1               &   73.5                  \\ \cline{2-8} 
 & C     & GT Attribute Known   & weighted    & Direct          & $\times$1     &   89.3               &   73.8                \\ \cline{2-8} 
 & D     & SSD-LLM Attribute    & random      & Direct          & $\times$1     &   89.5               &   76.2                   \\ \cline{2-8} 
 & E     & SSD-LLM Attribute    & weighted    & Direct          & $\times$1     &   89.8               &   77.4                   \\ \cline{2-8} 
 & F     & SSD-LLM Attribute    & weighted    & LLM Sentence    & $\times$1     &   90.1               &   78.3                   \\ \cline{2-8} 
 & G     & SSD-LLM \& LLM Suggest  & weighted    & LLM Sentence    & $\times$1     &   90.5               &   79.1                   \\ \cline{2-8} 
 & H     & SSD-LLM Attribute    & weighted    & LLM Sentence    & $\times$2     &   91.1               &   79.2                   \\ \hline
\end{tabular}
}
\caption{Ablation study on the sample mode and SD prompt mode.}
\label{ablation_diffusion}
\end{table}

\section{Discussion}
Although our method, SSD-LLM, has already shown effectiveness in various settings, the current exploration of the algorithm is limited to image datasets. Besides, this approach may bear the potential bias from the MLLMs and LLMs. However, it may be reduced from extra human-in-the-loop guidiances if available.

For future works, we suggest the following promising directions:
\begin{itemize}
    \item \textbf{Structure Format} The four-layer subpopulation structure can be expanded to more suitatble structures according to specific task requirements.
    \item \textbf{Downstream Tasks} SSD-LLM can have more applications in various computer vision and multimodality tasks, e.g. object detection and VQA.
    \item \textbf{Dataset Construction} The subpopulation structure obtained from SSD-LLM holds the potential to guide dataset construction with better fairness~\cite{wang2022revise} or further supporting the construction of unbiased datasets~\cite{liu2024decadesbattledatasetbias}.
    \item \textbf{Technical Extensions} The core procedures of SSD-LLM, using LLM to conduct group-level summarizations, can be extended to more types of contents including patterns of model hallucinations.
\end{itemize}

\section{Conclusion}

In this work, we present the first systematic exploration of subpopulation structure discovery. 
We provide a precise definition of subpopulation structure and introduce a fine-grained criteria to determine the structures. 
We propose SSD-LLM for automatic subpopulation structure discovery incorporating elaborate prompt engineering techniques .
SSD-LLM  can be combined with subsequent operations to better tackle several downstream tasks, including dataset subpopulation organization, subpopulation shift and slice discovery, with minor Task-specific Tuning.
We conduct extensive experiments to verify our proposed methods, demonstrating the remarkable effectiveness and and generality of SSD-LLM.

\section*{Acknowledgements}
Shanghang Zhang is supported by the National Science and Technology Major Project of China (No. 2022ZD0117801).


%
%
\bibliographystyle{splncs04}
\bibliography{main}
\end{document}